\pdfoutput=1

\documentclass[11pt]{article}

\usepackage[]{acl}
\usepackage{caption}
\usepackage{subcaption}
\usepackage{times}
\usepackage{latexsym}
\usepackage{graphicx}
\usepackage{todonotes}
\usepackage{booktabs}
\usepackage{amsmath}
\usepackage{tabularx}
\usepackage[T1]{fontenc}

\usepackage[utf8]{inputenc}

\usepackage{microtype}

\usepackage{inconsolata}

\title{MovieSum: An Abstractive Summarization Dataset for Movie Screenplays}

\author{Rohit Saxena \qquad Frank Keller \\
 Institute for Language, Cognition and Computation\\
 School of Informatics, University of Edinburg \\
 10 Crichton Street, Edinburgh EH8 9AB \\
  \texttt{rohit.saxena@ed.ac.uk} \quad \texttt{keller@inf.ed.ac.uk}}

\begin{document}
\maketitle
\begin{abstract}
Movie screenplay summarization is challenging, as it requires an understanding of long input contexts and various elements unique to movies. Large language models have shown significant advancements in document summarization, but they often struggle with processing long input contexts. Furthermore, while television transcripts have received attention in recent studies, movie screenplay summarization remains underexplored. To stimulate research in this area, we present a new dataset, MovieSum,\footnote{Our dataset and code is available at \url{https://github.com/saxenarohit/MovieSum}.} for abstractive summarization of movie screenplays. This dataset comprises 2200 movie screenplays accompanied by their Wikipedia plot summaries. We manually formatted the movie screenplays to represent their structural elements. Compared to existing datasets, MovieSum possesses several distinctive features: (1)~It includes movie screenplays, which are longer than scripts of TV episodes. (2)~It is twice the size of previous movie screenplay datasets. (3)~It provides metadata with IMDb IDs to facilitate access to additional external knowledge. We also show the results of recently released large language models applied to summarization on our dataset to provide a detailed baseline.

\end{abstract}

\section{Introduction}

Large language models have shown significant improvements in abstractive summarization in recent years \citep{zhong2022dialoglm,10.1145/3404835.3462846,zhong-etal-2021-qmsum,zhang-etal-2022-summn}, aiming to produce a concise and coherent summary of the input document. However, these models often struggle when the input context is long, particularly when the relevant information is distributed across the document \citep{liu2023lost}. To better understand this phenomenon and to advance research, datasets are needed that not only contain long-form documents but also have the property that important information is dispersed throughout the document. Movie screenplays have these characteristics: to generate a faithful summary, an understanding of characters and events across the entire length of the screenplay is required.

More recently, narrative summarization research has focused on TV shows and books \citep{kryscinski2021booksum,moskvichev-mai-2023-narrativexl}, with less attention given to movie screenplays \citep{gorinski-lapata-2015-movie,papalampidi-etal-2020-screenplay}. Notably, \citet{chen-etal-2022-summscreen} introduced a dataset of TV show transcripts which has gained considerable interest and was included in a long document summarization benchmark \citep{shaham-etal-2022-scrolls}. But unlike movie screenplays, TV episode transcripts tend to be relatively short and predominantly comprise spoken dialogue with minimal scene or character descriptions. Additionally, they are not self-contained, as the events or characters from previous episodes can be referred to. In contrast, movie screenplays are structured documents with various screenplay elements such as scene headings, locations, character names, dialogues and detailed scene descriptions. These are written by screenwriters and are characteristically formatted to denote each element. 

The largest current movie screenplay dataset \citep{gorinski-lapata-2015-movie,gorinski-lapata-2018-whats} comprises 917 automatically formatted screenplays (\mbox{ScriptBase-j}), with the most recent movie from 2013. We built MovieSum, a new movie screenplay dataset for abstractive summarization, which consists of 2200 movies, more than twice the size \mbox{ScriptBase-j}. Importantly, our new dataset has been formatted using a professional script writing tool and paired with Wikipedia plot summaries. Each movie is also tagged with its IMDB IDs to facilitate the collection of other external knowledge in rhw future. The dataset consists of movies spanning a wide range of genres from 1930 to~2023. 

We provide a detailed description of MovieSum, including the steps for collecting and filtering screenplays and statistics and comparison with other narrative datasets. We conduct extensive experiments to evaluate the performance of state-of-the-art summarization models on MovieSum, demonstrating its utility as a benchmark dataset for narrative summarization research. Experiments suggest that recent models struggle with long abstractive summarization, and we hope that our efforts will inspire further research in this area. Furthermore, we provide qualitative analyses of how the structure of the screenplay can be utilized in generating summaries.

\begin{figure*}[ht]
  \centering
  \begin{subfigure}[b]{0.4\textwidth}
  \centering
\includegraphics[width=\textwidth]{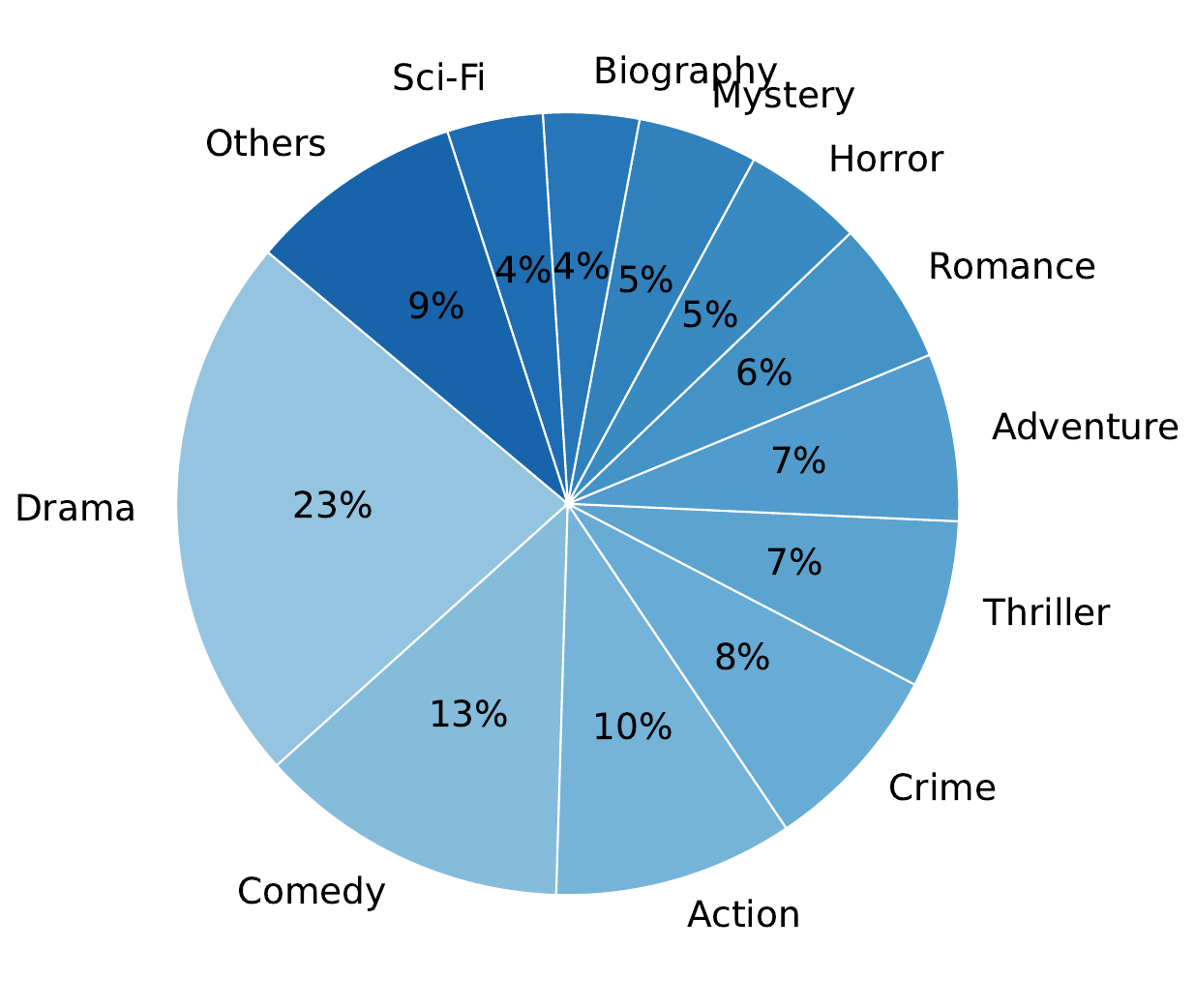}
\caption{Genres Distribution.}
\label{fig:genre_dist}
\end{subfigure}
 \hfill
  \begin{subfigure}[b]{0.5\textwidth}
\includegraphics[width=\textwidth]{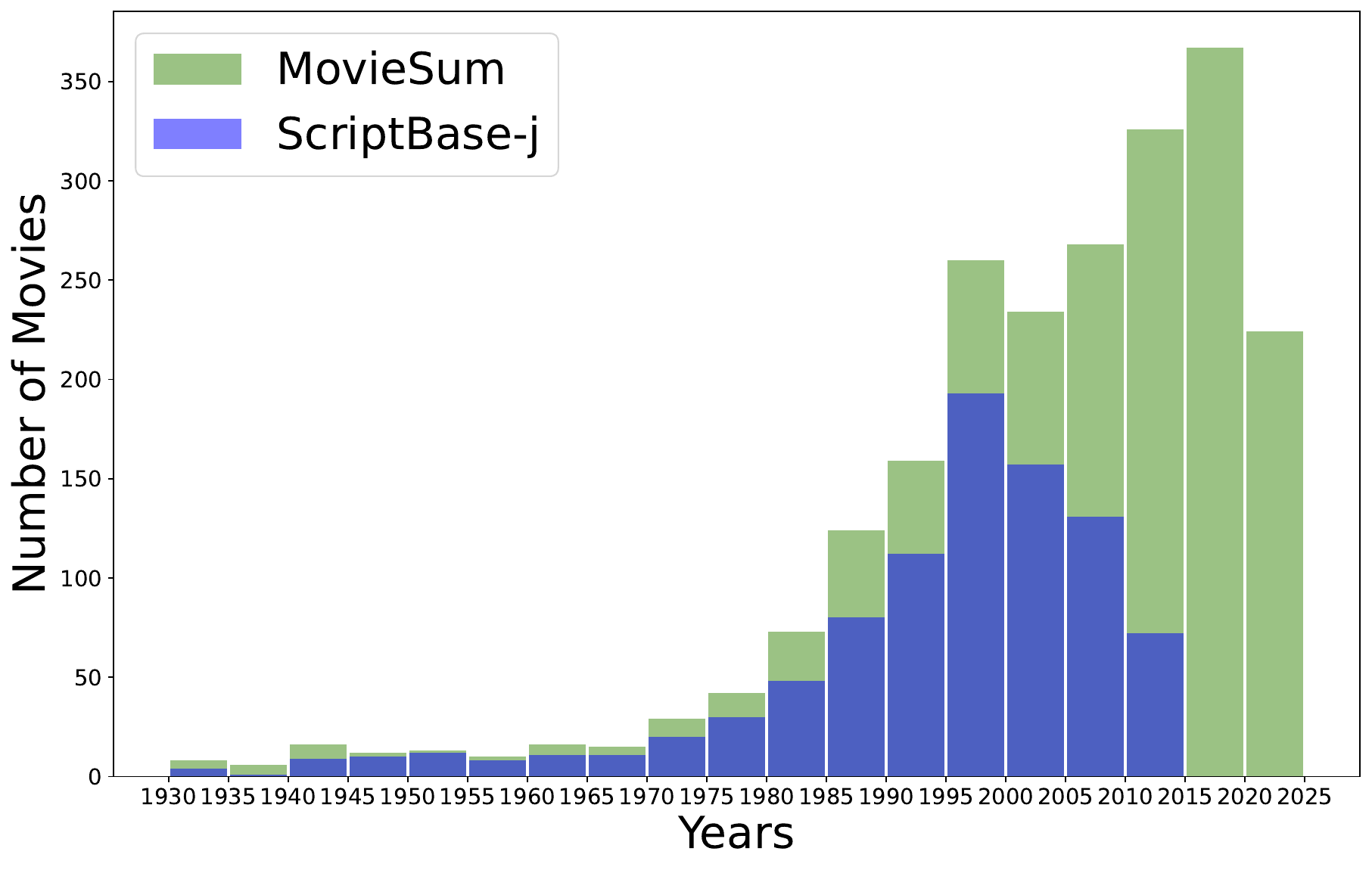}
\caption{Release years compared to ScripBase-j}
 \label{fig:year_dist}
 \end{subfigure}
  \caption{Distribution of movie genres and release years in the dataset.}
  \label{fig:fig1}
  
\end{figure*}
\section{The MovieSum Dataset}
We present MovieSum, a movie screenplay abstractive summarization dataset that consists of 2200 movie screenplay-summary pairs. All movie screenplays in the dataset are in English. 

\subsection{Collection of Movie Screenplays} 
We collected movie screenplays from a range of movie screenplay websites.\footnote{\url{https://www.scriptslug.com/}, \url{https://imsdb.com/}, \url{https://www.dailyscript.com/}} In total, we assembled 5,639 movie screenplays documents in various text format along with metadata of movie name, IMDB identifier, and release year. If the IMDB identifier was missing, we extracted it using the IMDB database. We then manually removed movies based on two criteria. Firstly, we removed any duplicate movie screenplays by using the movie names and release years to identify the duplicates. Secondly, we filtered out screenplays which did not have text or were incomplete.

\subsection{Screenplay Formatting}
Movie screenplays are structured documents with various script elements such as scene headings (also known as slug lines), characters' names, dialogues, and scene descriptions (or actions). These elements have specific markers based on spacing. Most of this formatting is lost when extracting text from these movie screenplay documents, making it challenging to retrieve the elements using regular expressions. 
To ensure the quality of the dataset, after filtering,  we manually corrected the movie screenplay and formatted each movie screenplay using Celtx,\footnote{\url{https://www.celtx.com/}} a professional screenplay writing tool. This process preserved the format of movie screenplay elements. We further filtered out screenplays with encoding or optical character recognition errors.

\subsection{Collection of Wikipedia Plot Summaries}

To build a robust summarization dataset, it is necessary to collect high-quality human-written summaries. Similar to previous work \citep{kocisky-etal-2018-narrativeqa}, we collected Wikipedia plot summaries, which we found to be of high quality, helped by the fact that Wikipedia summaries follow a consistent set of guidelines for movie plot summaries.\footnote{\url{https://en.wikipedia.org/wiki/Wikipedia:How_to_write_a_plot_summary}}

To collect the Wikipedia plot summary, we first extracted the Wikipedia page of the movie using the movie name and year, then collected text under the Plot section. We filtered out movies where the Wikipedia page or the plot section was unavailable. 

This process resulted in 2200 manually formatted movie screenplays with corresponding Wikipedia summaries.

\begin{table}[htbp]
\centering
\scalebox{0.9}{%
\begin{tabular}{@{}cccc@{}}
\toprule
\multicolumn{4}{c}{\% Novel n-grams in Summary} \\
1-grams   & 2-grams   & 3-grams   & 4-grams \\
\midrule
31.69     & 68.88    & 93.12 & 98.6 \\
\bottomrule
\end{tabular}}
\caption{Statistics for percentage of novel n-grams in the MovieSum summaries.}
\label{tab:ngram}
\end{table}

\begin{table*}[tb]
\centering
\scalebox{0.8}{%
\begin{tabular}{@{}lllccccc@{}}
\toprule
Datasets & Domain & Type & Task &  Size & Doc. Len. & Sum. Len \\
\midrule
ScriptBase--j & Movies & Formatted Screenplays  & Summarization & 917& 29K & 753\\
ScriptBase--alpha & Movies & Formatted/Raw Screenplays  &  Summarization&  917 / 359 & 29K & 738\\
SummScreenFD & TV Episodes & Raw Transcripts  & Summarization& 4348& 7605&113 \\
NarraSum & Movies/TV & Plot Summary & Summarization &122K &786 &147 \\
NarrativeXL & Books & Raw Text & QA & 1500 &87K & -- \\
NarrativeQA & Books/Movies & Raw Text/Screenplays   & QA/Summarization & 783 / 789 & 61K & 650 \\
BookSum & Books & Raw Text   & Summarization & 405 & 112K & 1167 \\
\midrule
\textbf{MovieSum} & Movies& Formatted Screenplays & Summarization& 2200& 29K & 717 \\
\bottomrule
\end{tabular}
}
\caption{Comparison of MovieSum with different narrative datasets for movies and TV shows.}
\label{tab:dataset_comp}
\end{table*}

\section{Dataset Analysis}

This results in a dataset consisting of 2200 manually formatted movie screenplays along with their corresponding summaries. The average length of the screenplays is 29k words, with an average summary length of 717 words. Importantly, this dataset is twice the size of the previously available movie screenplays dataset with formatted movie screenplays \cite{gorinski-lapata-2015-movie}. Figure~\ref{fig:genre_dist}. illustrates the genre distribution of the movies within the dataset and showcases the broad range of genres. In Figure~\ref{fig:year_dist}, the distribution of release years is depicted, revealing that the movies span a wide range of years, with a substantial number of them originating in recent years.

To study the abstractiveness of the summary, we report the percentage of novel n-gram in Table \ref{tab:ngram} as reported by \citet{10.1162/tacl_a_00373,zhao-etal-2022-narrasum}. It shows that a high number of 3-gram and 4-gram are novel in summary and not present in the movie screenplays implying high abstractiveness of the summary. See Appendix~\ref{sec:add_stats} for further analysis on the abstractiveness of the summary.

\begin{table*}[tb]
\centering
 \scalebox{0.9}{%
\begin{tabular}{@{}lrrrrrr@{}} 
\toprule            
 & \textbf{R-1} & \textbf{R-2} & \textbf{R-L} & \textbf{BS\textsubscript{p}} & \textbf{BS\textsubscript{r}} & \textbf{BS\textsubscript{f1}} \\ 
\midrule
Lead-512       & 10.35 & 1.27 & 9.84 & 49.25 & 43.59 & 46.23     \\
Lead-768       & 14.43 & 1.79 & 13.76 & 49.29 & 45.7 & 47.41        \\
Lead-1024       & 17.93 & 2.24 & 17.15 & 49.12 & 46.91 & 47.98      \\
TextRank & 33.32 & 5.27 & 32.10 & 51.46 & 52.47 & 51.85 \\
\midrule
FLAN-UL2 8K (ZS) & 23.62 & 4.29 & 22.01 & 52.9 & 49.57 & 50.87\\
Vicuna 13B 16K (ZS)  & 16.35 & 3.55 & 15.44 & 48.89 & 48.49 & 47.07\\
TextRank with Vicuna 13B 16K (ZS) &	17.14	& 3.68 &	15.47 &	59.24 &	49.05 & 53.57 \\	
Moving Window Vicuna 13B 16K (ZS) &	19.56 &	3.32 &	18.57 &	54.95 &	48.7 &	51.53 \\	

\midrule
Pegasus-X 16K  &  42.42 &  8.16 &  40.63 & 58.81 & 50.56 & 54.36 \\
LongT5 16K & 41.49 & 8.54 & 39.78 & 56.09 & 55.36 & 55.68\\
Longformer (LED) 16K & 44.85 & 9.83 & 43.12 & 59.11 & 58.43 & 58.73 \\
\midrule
Dialogues Only (LED) & 44.68 & 10.02 & 42.94 & 59.30 & 58.29 & 58.74 \\
Description Only (LED) & 44.72 & 9.72 & 42.92 & 59.47 & 58.45 & 58.92 \\
Heuristic Only (LED) & 44.45 & 9.78 & 42.71 & 58.93 & 58.15 & 58.54\\

\bottomrule
\end{tabular}%
 }
\caption{Results of summarization models on the MovieSum dataset. We report ROUGE and BERTScores for the baselines and other summarization models.}
\label{tab:summ_result}
\end{table*}

\subsection{Comparison with Existing Datasets}

We compare our dataset with various datasets in the narrative domain, and the statistics are reported in Table~\ref{tab:dataset_comp}. These datasets include \mbox{ScriptBase-j} \citep{gorinski-lapata-2015-movie}, ScriptBase-alpha \citep{gorinski-lapata-2015-movie}, SummScreenFD \citep{chen-etal-2022-summscreen}, NarraSum \citep{zhao-etal-2022-narrasum}, NarrativeXL \citep{moskvichev-mai-2023-narrativexl}, NarrativeQA \citep{kocisky-etal-2018-narrativeqa}, and BookSum \citep{kryscinski2021booksum}. Notably, BookSum and NarrativeXL have a longer average document length but comprise books, not screenplays. SummScreenFD consists of TV show episode transcripts, which are much shorter in both document and summary length. Importantly, SummScreen consists of community-contributed transcripts and primarily comprises dialogues, unlike screenplays, which include detailed scene descriptions. Also, TV show episodes are not self-contained, as events or characters from previous episodes can be referenced.  NarraSum contains plot summaries as documents rather than actual screenplays, and therefore has the lowest average document length among the datasets we compare. NarrativeQA includes both books and movie screenplays, with books being notably lengthy, making the average document length comparable to book datasets. However, it consists of only 789 unformatted movie screenplays.

Both \mbox{ScriptBase-j} and ScriptBase-alpha datasets are close to our screenplay dataset. \mbox{ScriptBase-j} contains formatted screenplays, whereas ScriptBase-alpha comprises the unformatted raw text of screenplays. It is important to note that \mbox{ScriptBase-j} is a subset of ScriptBase-alpha, which consists of 917 formatted screenplays. On the other hand, ScriptBase-alpha includes an additional 359 movies. Our work can be considered as the extension of \mbox{\mbox{ScriptBase-j}} as it also consists of formatted screenplays. At the same time, we overcome two critical limitations of SciptBase-j:

(1) The formatting of the movie screenplay was performed automatically. Although it is easy to detect the scene heading based on rules and string matching, it is challenging to distinguish dialogues, character names, and scene descriptions. The work does not provide any details regarding the automatic formatting strategy. MovieSum, on the other hand, includes all the movies from \mbox{\mbox{ScriptBase-j}} formatted using professional screenplay tools.

(2) Both subsets of ScriptBase consist of movies released until 2013.  In contrast, MovieSum also includes recently released movies. This is crucial for ensuring that summarization models remain robust to new movie narrative conventions. The movie release years are graphed in Figure~\ref{fig:year_dist}.

\section{Experiments}\label{sec:Experiments}

We evaluate the MovieSum dataset using several baselines and state-of-the-art neural abstractive summarization models. We first report the \textbf{Lead-N} baseline, which simply outputs the first $N$ tokens of the movie script as the movie summary. We varied the value $N$ to understand the impact of summary length on performance and report results for \textbf{Lead-512}, \textbf{Lead-768}, and \textbf{Lead-1024}. For the extractive baseline, we used \textbf{TextRank} \citep{mihalcea-tarau-2004-textrank}, a graph-based unsupervised extractive summarization method. For instruction-tuned large language models, we used \textbf{Vicuna 1.5 13B} 16K \citep{zheng2023judging}, built on Llama-2 \citep{touvron2023llama}, and \textbf{FLAN-UL2} \citep{tay2023ul,wei2022finetuned} in a zero-shot setting. For fine-tuned models with long inputs, we utilized \textbf{LongT5} \citep{guo-etal-2022-longt5}, \textbf{PEGASUS-X} \citep{phang-etal-2023-investigating}, and the Longformer Encoder-Decoder (\textbf{LED}) model \citep{Beltagy2020Longformer}. We fully fine-tuned these models and report results on the test set. The implementation details of the models are mentioned in Appendix~\ref{sec:impl_det}.
 
\subsection{Results}
Table~\ref{tab:summ_result} shows the summarization evaluation results using ROUGE F1 (1/2/L) scores \citep{lin-2004-rouge} and BERTScore \citep{zhang2019bertscore} on MovieSum. The Lead baseline performs better with a higher number of words, achieving the best result with 1024 words. This is not surprising, as the ROUGE metric is known to give higher scores for longer summaries \citep{schluter-2017-limits}. In the case of zero-shot, FLAN-UL2 8K performed substantially better than Vicuna 13B 16K. This confirms that a longer context does not necessarily lead to attention to the full context length. We also tested reducing the input context using TextRank, which only marginally improves the zero-shot performance for Vicuna 13 model. To further utilize the full context, we also report results for moving window chunk-based zero-shot summarization and concatenating the generated summaries. This performs better compared to using only the 16K context length. The best performance was achieved with fine-tuned models with longer context lengths. Pegasus-X, LongT5, and LED perform similarly well, with LED demonstrating superior performance.

\section{Analysis of Screenplay Structure}
We analyze the importance of screenplay elements, dialogue and scene description, and their impact on summarization performance. We selected the best model from the full fine-tuned experiment and studied the effects of fine-tuning solely on dialogues and descriptions. Additionally, we investigated the effects of selectively using both dialogue and scene description based on heuristics similar to \citet{pu-etal-2022-two}. Results in Table \ref{tab:summ_result} show a marginal impact on summarization when dialogues or scene descriptions are removed. We also found the heuristic based removal of input context method results is comparable to the full-text results. This suggests that the current model does not fully utilize the inherent structure of the document, and new methods for content selection and summarization should consider both screenplay elements for movie script summarization.

\section{Discussion and Conclusion}

We introduce the MovieSum dataset, comprising formatted and recent movie screenplays paired with Wikipedia summaries. Our experiments demonstrate that it is a challenging dataset even for a large language model with a long input length. We hope that MovieSum will enable future research in the area of movie screenplay understanding and abstractive summarization.

\section*{Limitations}
Limitations of the work include that the dataset consists of movie screenplays and their corresponding summaries only in English. Models trained on MovieSum may not generalize well to multilingual summarization tasks or applications requiring cross-lingual understanding.

\section*{Ethics Statement}
\paragraph{Large Language Models:} This paper uses pre-trained large language models, which have been shown to be subject to a variety of biases, to occasionally generate toxic language, and to hallucinate content. Therefore, the summaries generated using our dataset should not be released without automatic filtering or manual checking.
\paragraph{Bias:} Despite efforts to include a wide range of movies, the dataset may not fully represent the diversity of cinematic styles, languages, or cultural contexts. Models trained on MovieSum may therefore exhibit biases towards the types of movies included.

\section*{Acknowledgements}
This work was supported in part by the UKRI Centre for Doctoral Training in Natural Language Processing, funded by UK Research and Innovation (grant EP/S022481/1), Huawei, and the School of Informatics at the University of Edinburgh. We would like to thank the anonymous reviewers for their helpful feedback.

\bibliography{acl_latex}

\appendix
\section{Implementation Details}
\label{sec:impl_det}
For TextRank, we set the parameter $words=1024$. We randomly split the dataset into 1800/200/200 as a train/val/test set to train the models. We used the base variants of Pegasus-X, LongT5, and LED for fine-tuning. Each input sequence for the movie is truncated to 16,384 tokens (including special tokens) to fit into the maximum input length of the model. We used AdamW as an optimizer ($\beta_1 = 0.9$, $\beta_2 = 0.99$). For LED and LongT5, we used a learning rate of 2e-5 with a cosine scheduler and a warmup ratio of 0.01. We set the $max\_new\_token$ to 1024 with greedy decoding for all the experiments. For Pegasus-X, we found that a learning rate of 5e-5 performed better with a linear warmup strategy and a warmup ratio of 0.01.  All models were trained for 50 epochs, and the best model was selected using the ROUGE-1 on the validation set.  The rest of the configurations for the models were kept as default. All the models were trained on A100 GPU with 80GB memory. We used the Huggingface $evaluate$ library for the implementation of the metrics.

\begin{figure}[ht]
  \centering
\includegraphics[width=0.8\columnwidth]{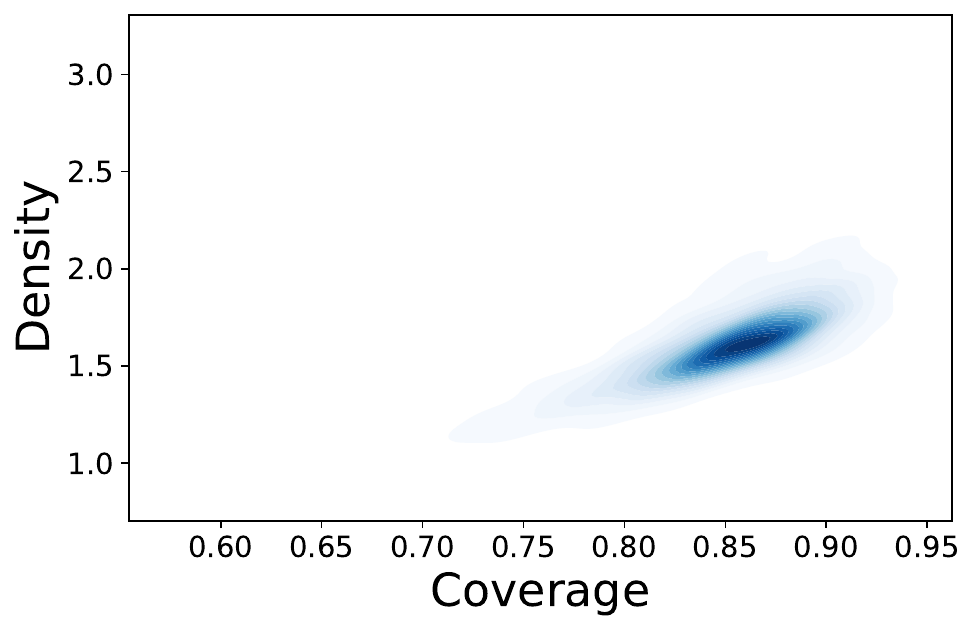}
\vspace{-2ex}
  \caption{Coverage-Density plot of the summaries.}
  \label{fig:density_plot}
\end{figure}

\section{Additional Statistics of Dataset}
\label{sec:add_stats}
To further understand the abstractiveness of the summaries we computed the coverage and density of the summaries as discussed by \citet{10.1162/tacl_a_00373}. The low density in Figure \ref{fig:density_plot}, indicates low overlap between the summary and the screenplays. 

\section{Prompt Template}
\label{sec:promt_temp}
For the zero-shot experiments in Section \ref{sec:Experiments}, we used the following prompt template:

\vspace{1em} 
\noindent\textbf{Prompt:} Summarize the following movie script. \\
Movie Script: \{movie script text\} \\
Summary:
\vspace{1em} 

\begin{figure}[htbp]
  \centering
\includegraphics[width=0.8\columnwidth]{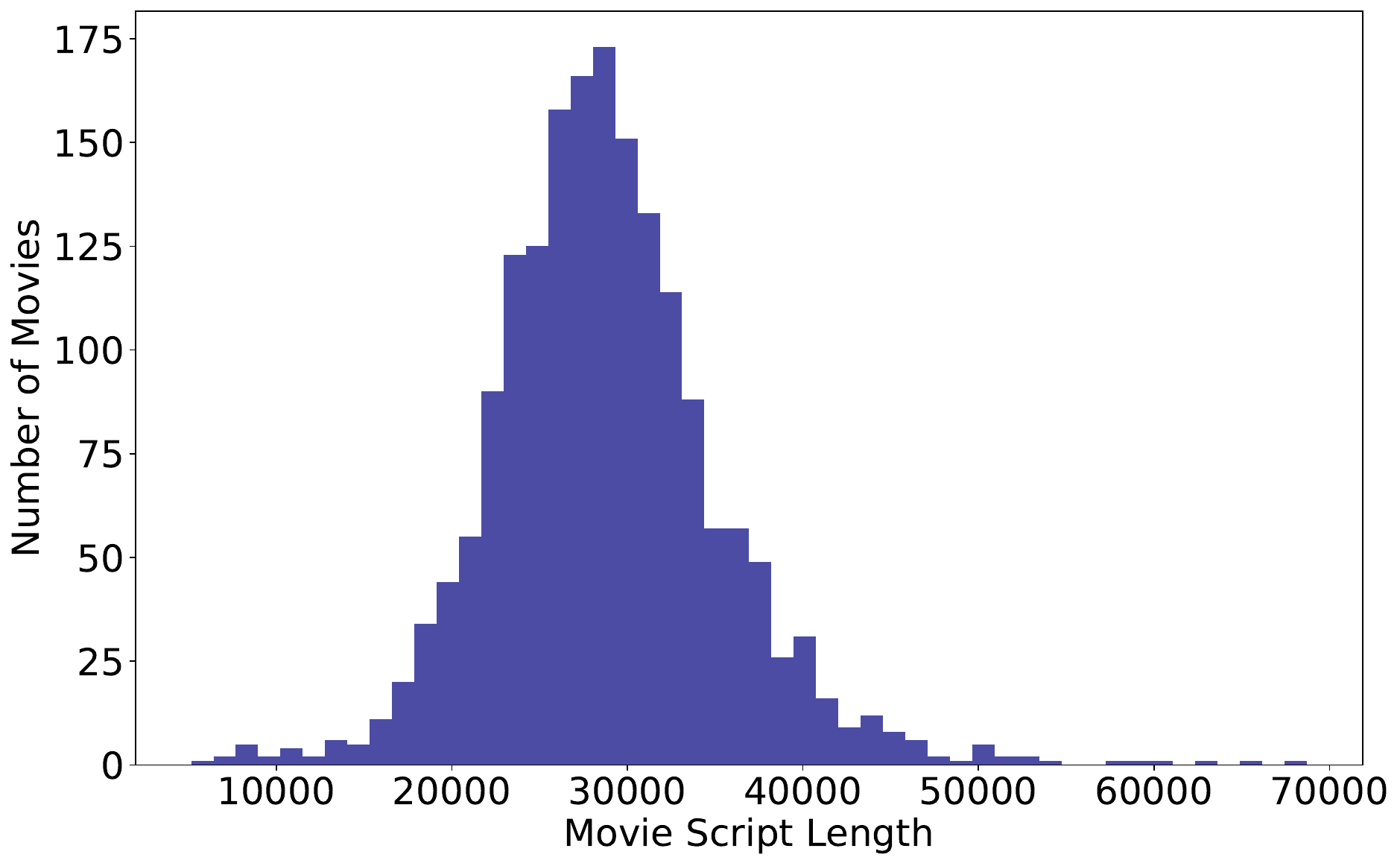}
\vspace{-2ex}
  \caption{Distribution of movie script length from the training set.}
  \label{fig:script_length_dist}
\end{figure}

\begin{figure}[htbp]
  \centering
\includegraphics[width=0.8\columnwidth]{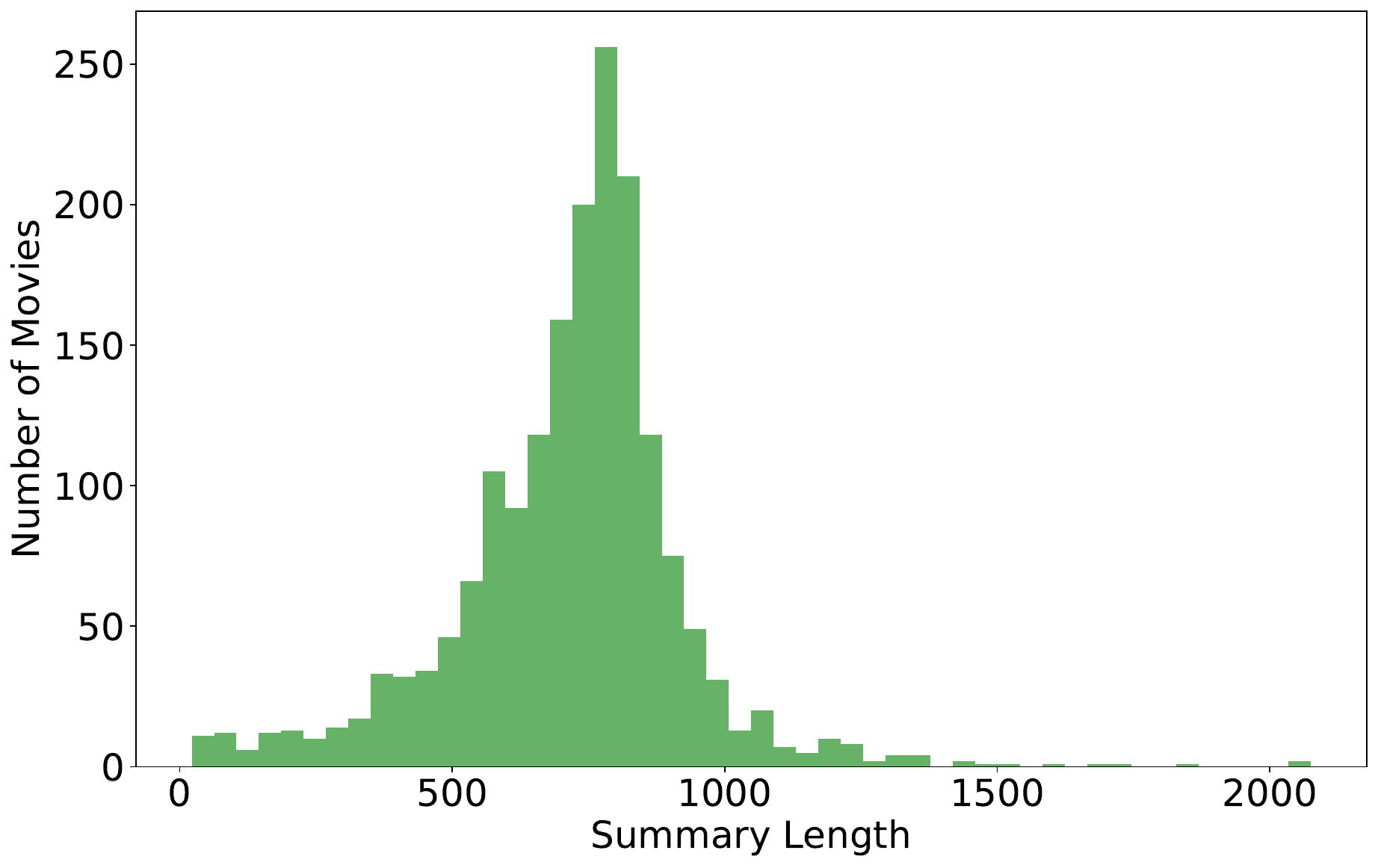}
\vspace{-2ex}
  \caption{Distribution of summary length from the training set.}
  \label{fig:summary_length_dist}
\end{figure}

\section{Length Distribution}
Figure~\ref{fig:script_length_dist} and \ref{fig:summary_length_dist} show the length distribution for movie scripts and their summaries across the training set. The mean length of movie scripts is 29K words, and the average length of summaries is 714 words.

\section{Example of a Movie Screenplay}
\label{sec:example_movie}
\begin{figure*}[ht]
  \centering
\includegraphics[width=\textwidth]{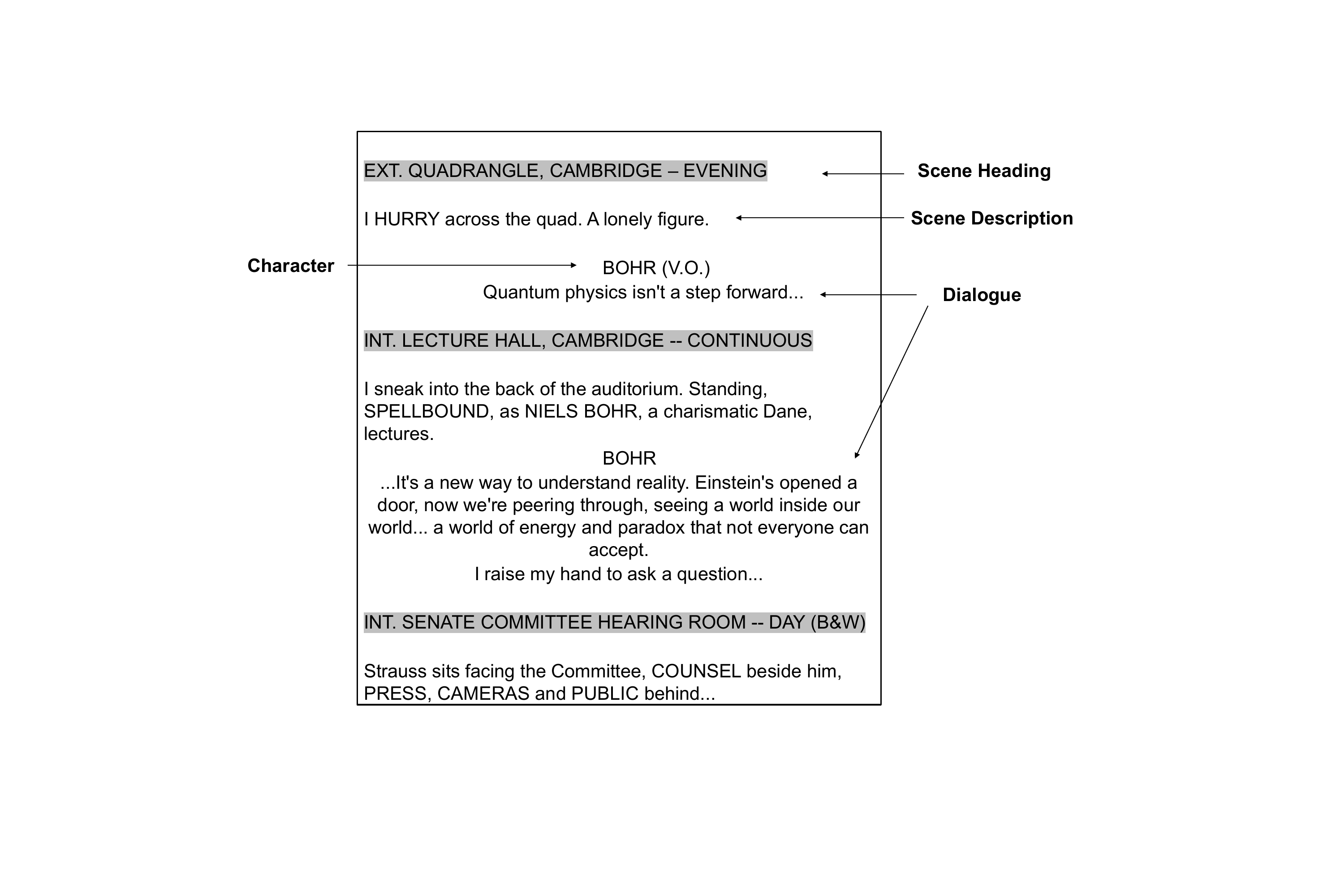}
 \vspace{-10ex}
  \caption{Example of cleanly formatted scenes from a movie screenplay.}
  \label{fig:screenplay}
\end{figure*}

Figure~\ref{fig:screenplay} shows an example of a cleanly formatted screenplay with distinct elements such as scene heading, characters, and dialogues. All the files are converted into XML using Celtx tool.

\section{Sample of Movie Summary}\label{sec:generatedsumm}

Table~\ref{tab:gen_summ} shows sample of generated summary of a movie using fine-tuned LED model (full-text).

\begin{table*}[!htbp]
\small
\begin{tabularx}{\linewidth}{X}
\toprule
\textbf{Movie: Oppenheimer}\\
\midrule
\textbf{Generated Summary}\\
 In 1947, American physicist Dr. Robert Oppenheimer (Samuel L. Jackson) is appointed chairman of the American Enterprise Institute. After giving a lecture on quantum mechanics at Columbia University, Oppenheimer asks him, "How did you come to the place you are?" Angered by Oppenheimer's failure to present convincing proof of his theory, Oppenheimer dismisses him. Eltenton moves to California to start a school of his own, and he teaches himself the art of quantum mechanics. Lewis Strauss (Nicholson), a young scientist from New York City and a fellow associate at the institute, meets fellow physicist Dr. Robert Oppenheimer at a party meeting. They begin a relationship, and he introduces her to the Tolmans (Lerner) and their daughter, Peter. They form a secret organization, the Organization of American Unity, led by Dr. Robert Oppenheimer. Eventually, the Organization of American Unity is shut down. In 1958, Oppenheimer becomes the new chairman of the AEC, overseeing the construction of the Manhattan Engineer District. The AEC chairman, General Groves (Harvey Keitel), promotes Oppenheimer to general and orders him to head the school; he reluctantly does so. In 1964, Oppenheimer agrees to build a secret atomic weapon, the "Thermostellar Trigger" (Thermostellar Trigger) at Los Alamos, California. Oppenheimer asks the head of the school to help him construct a machine that could split the atom, but he is rebuffed. Angered that Oppenheimer would abandon the project, Oppenheimer urges him to leave the school and take a job in Washington, D.C. as a "secret member of the Communist Party". He argues that such a machine would be "destroyer of worlds". At the AEC meeting, he is overcome by guilt and despair, and Oppenheimer announces that he will start the school. In 1970, Oppenheimer died of a heart attack while trying to explain the concept of a supernova. Groves decides to start a secret organization, the Friends of the ABC, to support the project. Lawrence Bender (Edith Atwater), an old friend of Oppenheimer's, explains that in 1947 and 1954, he and several other scientists worked on a project to create a machine that could split the atom, but the team discovered that the machine could not get enough power to make the machine work. They sent the machine to Mexico, where it was secretly carried out. In 1969, Oppenheimer died of a heart attack while in the laboratory. Thirty years later, in 1997, Oppenheimer has died of a rare illness at the age of 73. Lawrence still works at the institute. In 2002, Oppenheimer loses his job because of the explosion. Thirty years later, in 2003, Oppenheimer, whose wife of over 25 years has died of a heart attack while working at the institute. Lawrence, Frank, and Jackie travel to Mexico to honor his friend and colleague, Dr. Hagenheimer. In a speech at his home following the death of his brother, Lawrence claims that he is ready to do what Oppenheimer asked of him. In a post-credits scene, as the screen fades to black.\\

\bottomrule
\caption{Model generated summary of the movie Oppenheimer. The summary consists of faithfulness errors.}
\label{tab:gen_summ}
\end{tabularx}
\end{table*}

\end{document}